# Unlocking the Potential of Large Language Models in the Nuclear Industry with Synthetic Data

**Muhammad Anwar [1,2], Daniel Lau [1,2], Mishca de Costa [1,2], and Issam Hammad [2]**
[1] Data Analytics and AI, Digital Technology and Services, Ontario Power Generation, Pickering, Ontario, Canada
[2] Department of Engineering Mathematics and Internetworking, Faculty of Engineering, Dalhousie University, Halifax, Nova Scotia
muhammad.anwar@opg.com

## Abstract

The nuclear industry possesses a wealth of valuable information locked away in unstructured text data. This data, however, is not readily usable for advanced Large Language Model (LLM) applications that require clean, structured question-answer pairs for tasks like model training, fine-tuning, and evaluation. This paper explores how synthetic data generation can bridge this gap, enabling the development of robust LLMs for the nuclear domain. We discuss the challenges of data scarcity and privacy concerns inherent in the nuclear industry and how synthetic data provides a solution by transforming existing text data into usable Q&A pairs. This approach leverages LLMs to analyze text, extract key information, generate relevant questions, and evaluate the quality of the resulting synthetic dataset. By unlocking the potential of LLMs in the nuclear industry, synthetic data can pave the way for improved information retrieval, enhanced knowledge sharing, and more informed decision-making in this critical sector.

## 1. Introduction

The nuclear industry is inherently data intensive. Vast volumes of technical documents, regulatory reports, and operational logs contain valuable insights—yet much of this information remains locked away in unstructured text formats. These documents, rich in technical details and critical operational data, are not immediately usable by advanced AI systems. State-of-the-art Large Language Models (LLMs) require well-structured data, typically in the form of question–answer (QnA) pairs, for effective training, fine-tuning, and evaluation.

Synthetic data generation presents an innovative approach to bridge this gap. By leveraging the powerful analytical capabilities of LLMs, synthetic data techniques can convert unstructured nuclear text into structured QnA datasets. This transformation not only overcomes challenges of data scarcity and stringent privacy requirements but also enables the creation of scalable, domain-specific resources. Ultimately, this method paves the way for improved information retrieval, enhanced knowledge sharing, and more informed decision-making across the nuclear sector. By unlocking the latent potential within nuclear documentation, synthetic data can be a step towards transforming how we harness AI for safer and more efficient nuclear operations.

## 2. LLMs in the Nuclear Industry: Obstacles, Challenges, and the Promise of Synthetic Data

The potential impact of LLMs in the nuclear industry is profound. They can revolutionize information retrieval, knowledge management, and even decision support systems. However, realizing this potential is not without its obstacles.

### 2.1 Obstacles and Challenges

Several factors limit the effective deployment of LLMs within the nuclear domain. Data scarcity is a major challenge, as strict security protocols and privacy regulations severely restrict access to high-quality, structured nuclear data. Unlike other industries with large, annotated datasets readily available, the nuclear sector must contend with limited human-annotated data, making it difficult to train robust LLMs. Additionally, privacy concerns further complicate AI adoption. The sensitive and classified nature of nuclear data demands rigorous privacy protection, and even when relevant data exists, sharing or annotating it poses significant risks. This greatly hinders the development of AI systems that rely on access to such data.

Another critical barrier is the need for structured data. LLMs function optimally with structured inputs, such as clear question–answer (QnA) pairs, yet the nuclear industry's data is predominantly unstructured. Technical reports, operational logs, and regulatory documents often contain critical information in free-text formats, requiring domain expertise to accurately extract and organize relevant details. The challenge is not only in structuring the data but also in ensuring that key technical details are preserved without compromising security or integrity. These combined challenges create a significant obstacle to the effective use of LLMs in nuclear applications, where precision, reliability, and compliance are paramount.

### 2.2 Synthetic Data Generation: A Scalable, Privacy-Preserving Solution

Synthetic data generation emerges as a transformative solution to these obstacles by leveraging LLMs to analyze and reinterpret unstructured nuclear text, thereby creating structured datasets that are both scalable and privacy-preserving. One of the primary benefits of synthetic data generation is its ability to mitigate data scarcity. Using synthetic techniques, extensive QnA pairs can be generated from limited seed data, expanding the dataset without requiring access to additional sensitive information. This provides a robust alternative to traditional human-annotated datasets while ensuring scalability.

Beyond addressing data scarcity, synthetic data generation also enhances privacy protection. Since synthetic data is algorithmically generated rather than extracted directly from classified or sensitive sources, it reduces the risk of data breaches and ensures compliance with stringent nuclear industry regulations. This allows organizations to create high-quality training datasets without exposing proprietary or classified content.

Furthermore, creating structured data from unstructured sources is a core advantage of this approach. LLMs, through advanced processing techniques such as iterative prompt engineering and context-aware extraction, can convert complex nuclear texts into well-organized QnA formats. This process includes multiple quality evaluation steps, ensuring that the resulting data is both accurate and aligned with domain-specific knowledge.
Through a carefully designed methodology that incorporates iterative refinement and human-in-the-loop validation, synthetic data generation can produce high-quality, diverse datasets tailored specifically for the nuclear industry. This approach not only overcomes traditional limitations posed by data scarcity and privacy constraints but also unlocks new possibilities for deploying advanced AI models in nuclear safety, reactor operations, and regulatory compliance.

## 3. Related Work

The use of Large Language Models (LLMs) for synthetic data generation has become an essential approach to overcoming data scarcity, privacy concerns, and the high costs of human annotation. Research demonstrates that synthetic datasets can effectively augment training data, particularly in specialized domains such as healthcare, information retrieval, and nuclear AI applications [1]. However, ensuring the accuracy, diversity, and cost-effectiveness of synthetic data remains a challenge.

Liu et al. [2] emphasize best practices in synthetic data generation, highlighting the need for factual accuracy and dataset diversity. Chan et al. [3] propose a cost-effective framework for synthetic data generation, categorizing methods into Answer Augmentation, Question Rephrasing, and New Question Evolution. Their findings indicate that different strategies impact model performance based on data availability and computational constraints. Additionally, retrieval-augmented generation (RAG) has emerged as a promising technique for improving factual accuracy, as demonstrated by Zhu et al. [4], who introduce RAGEval, a framework for evaluating domain-specific synthetic datasets.

Synthetic data has also shown promise in domain-specific AI applications. Patel et al. [5] present DataDreamer, a tool for reproducible LLM-driven synthetic data workflows. Braga et al. [6] show that LLM-generated datasets can replace human-authored QnA pairs in personalized information retrieval, while Kang et al. [7] leverage synthetic clinical records to enhance depression prediction models, preserving patient privacy while improving model robustness.

Despite these advancements, challenges remain. Hallucination, where LLMs generate incorrect but plausible content, is a major concern, requiring rigorous validation techniques [2]. Bias propagation is another risk, as synthetic datasets can inherit biases from LLM training data, necessitating filtering and diversity-enhancing strategies. Additionally, while synthetic data reduces annotation costs, large-scale generation remains computationally expensive, necessitating optimized prompt engineering and cost-effective data generation strategies [3].

## 4. Synthetic Data Generation Process

This section outlines the process of generating synthetic question-answer pairs from unstructured text data in the nuclear domain, with a specific focus on CANDU reactor technology. The approach employs Large Language Models (LLMs) to convert raw text from the "Essential CANDU" textbook [8] into structured QnA pairs for downstream AI model training and evaluation tasks in the nuclear industry. The process comprises several key steps which are outlined in Figure 1 and explained in the paper’s sections.

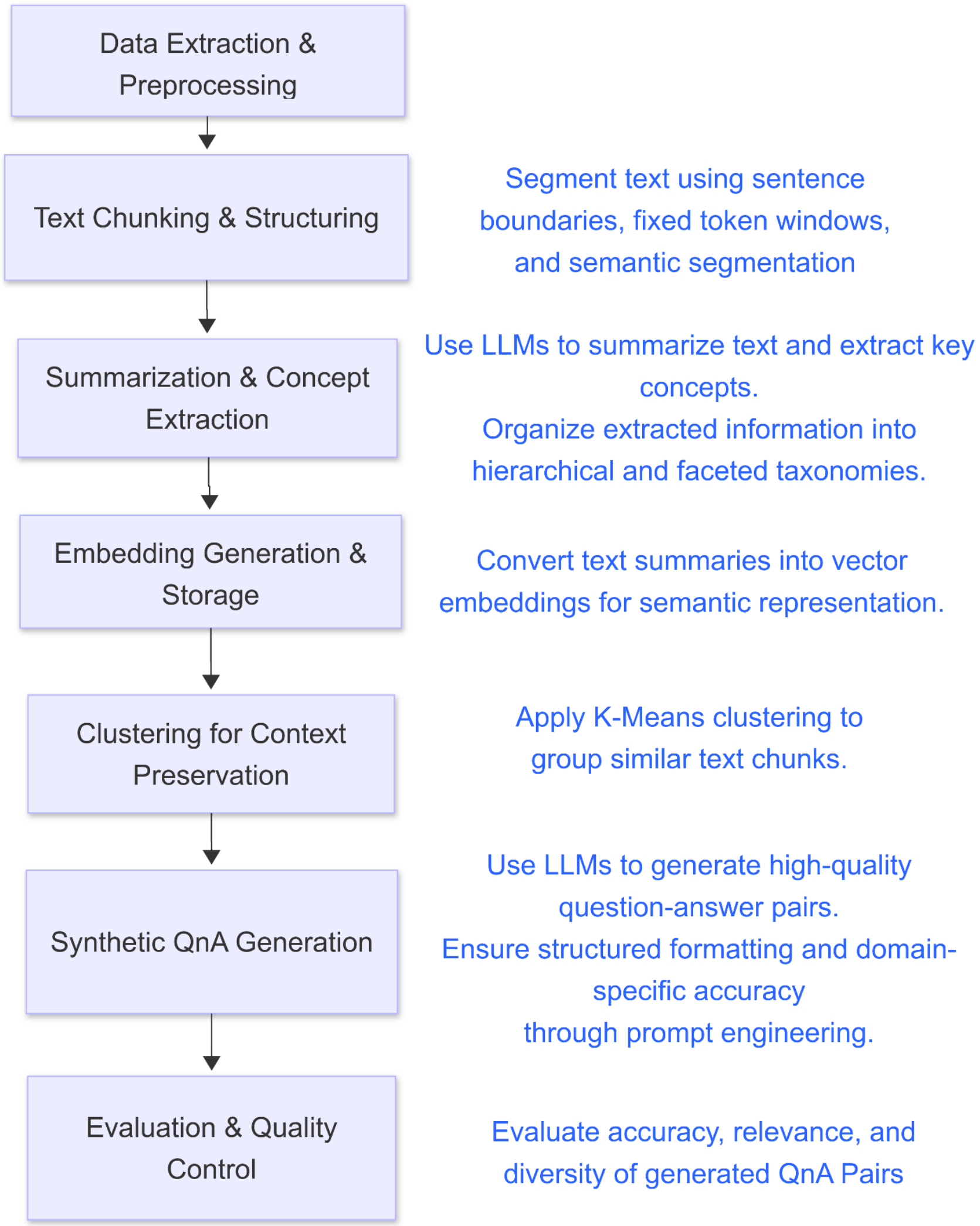

**Figure 1: Overall structure of synthetic QnA pairs pipeline**

## 4.1 Data Extraction and Preprocessing

### 4.1.1 Text Chunking Strategy

For this paper, since the CANDU textbook was used as the primary data source. The text was extracted using Azure Document Intelligence, which identifies meaningful separations within the textbook, such as paragraphs, new sections, and chapter divisions. These naturally occurring separations were leveraged to form chunks, which were further organized based on page numbers to maintain document structure and contextual integrity.

#### 4.1.2 Summarization for Key Concept Extraction

A careful constructed prompt is used to extract key information from the documents so that we can use it for clustering and downstream QnA generation. The structure of the summarization prompt is designed to guide the Large Language Model (LLM) towards extracting comprehensive and structured information from the technical text. By explicitly specifying the categories of information to extract (technical concepts, system components, operational processes, safety protocols, etc.), the prompt ensures that the LLM focuses on the most relevant aspects of the CANDU reactor technology. This structured approach not only facilitates the extraction of key insights but also ensures that the output is organized and formatted in a way that is conducive to further analysis and knowledge integration.

Furthermore, the prompt includes contextual information about the "Essential CANDU" textbook and its target audience. This helps the LLM to better understand the technical depth and scope of the text, leading to more accurate and relevant summarization. The emphasis on factual accuracy and clear organization in the prompt ensures that the extracted information is reliable and can be readily used for downstream tasks like clustering and QnA generation.

### 4.2 Embedding Generation and Clustering

This stage converts the textual data into a format suitable for efficient processing and analysis, preserving the contextual relationships within the text.

#### 4.2.1 Embedding Generation

To enable semantic-based analysis and retrieval, the textual chunks are transformed into numerical vector representations called embeddings. These embeddings capture the semantic meaning of the text, allowing for the comparison and grouping of similar chunks effectively.

The text-embedding-ada-002 model from Azure OpenAI is used for generating embeddings. This model produces 1536-dimensional vectors, providing a detailed representation of the text. These embeddings are important for tasks such as semantic search, context-aware question answering, and identifying similarities between different text segments.

#### 4.2.2 Clustering for Context Preservation

To maintain the contextual relationships between different text segments, semantically similar chunks are grouped using clustering techniques.

We utilize K-Means clustering to partition text embeddings based on their semantic similarity, ensuring that related chunks are grouped together for better downstream processing.

t-SNE (t-Distributed Stochastic Neighbor Embedding) is an effective technique for visualizing high-dimensional data in a lower-dimensional space while preserving local similarities. In this case, the t-SNE plot provides an intuitive way to observe the distribution of chunk embeddings and their respective clusters. The color-coded points correspond to different cluster assignments, revealing the structure of the dataset. The visualization indicates that the embeddings have been effectively grouped, with distinct regions corresponding to different clusters. This separation suggests that the clustering approach has successfully captured meaningful patterns in the text data, allowing for contextually relevant segmentation.

From the t-SNE plot as shown in Figure 2, it is evident that the clusters are well-separated, reinforcing the validity of the K-Means clustering approach. The clear distinctions between groups indicate that similar text segments are grouped together, supporting better semantic organization. The presence of minimal overlap between clusters suggests that the embeddings retain meaningful contextual relationships, which is crucial for downstream applications such as retrieval-augmented generation (RAG) and semantic search. This visualization confirms that the clustering strategy effectively maintains the integrity of contextual groupings within the synthetic data generation pipeline.

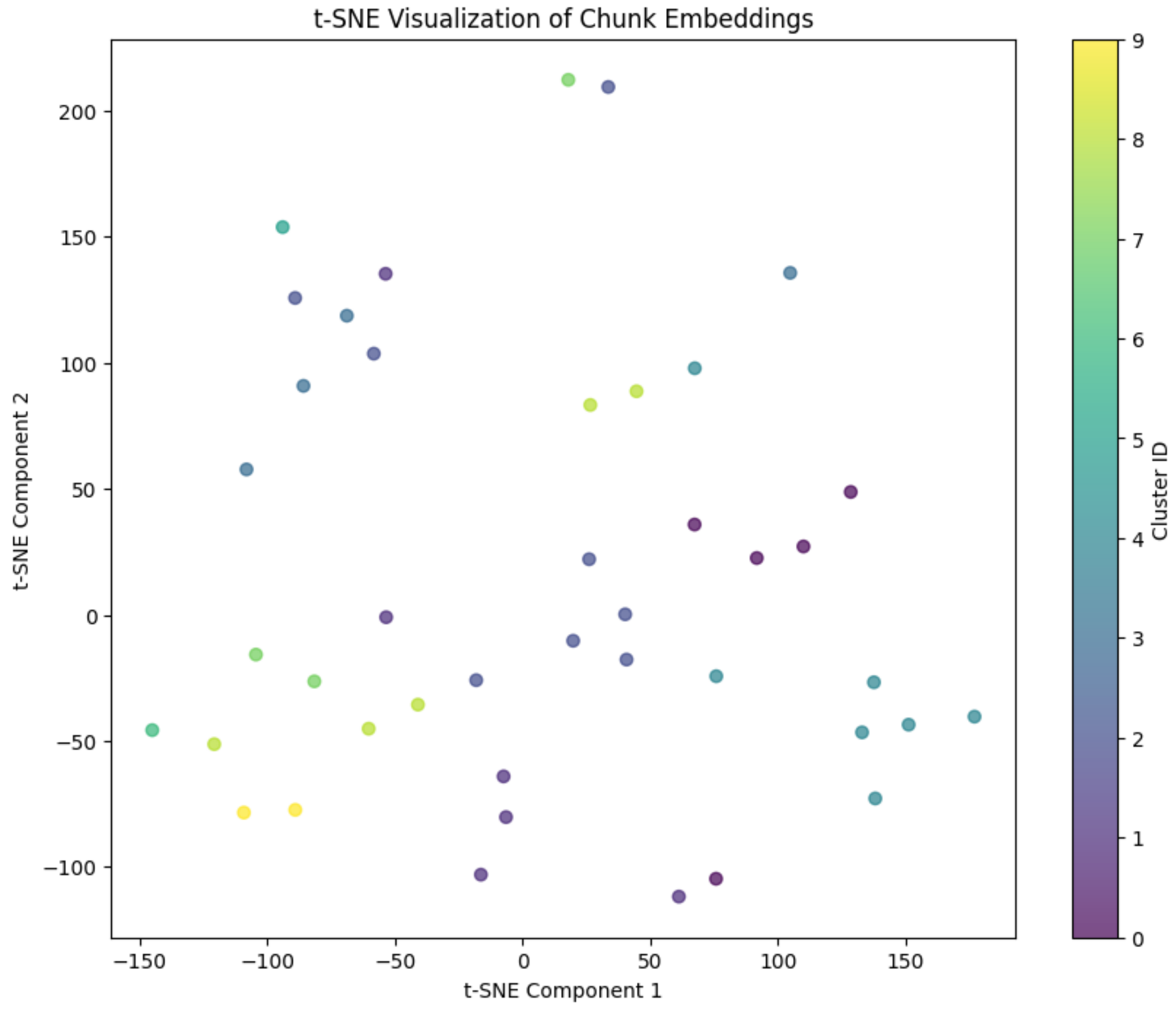

**Figure 2: t-SNE visualization of clusters**

### 4.3 Synthetic QnA Pair Generation

Prompt engineering was employed to develop question and answer pairs, ensuring the creation of a comprehensive and high-quality synthetic dataset for training and evaluating AI models in the nuclear engineering domain, specifically focusing on CANDU reactors. The QnA generation prompt was carefully designed with several key strategies in mind. First, contextualization and specificity were prioritized by explicitly referencing "The Essential CANDU" textbook, clearly defining the domain and purpose of the QnA pairs. This approach guides the LLM to generate questions and answers that are both relevant and technically accurate within the CANDU reactor context. Second, structured information and completeness were ensured by providing both the original text chunk and the extracted key information, allowing the LLM to access and integrate details from different parts of the text. This guarantees that the generated QnA pairs capture a complete and nuanced understanding of the concepts, even when the information is dispersed throughout the source material.

Another crucial aspect was diversity and complexity in the generated questions. The prompt was designed to encourage a broad range of question types, from fundamental recall to open-ended critical thinking questions, covering a wide spectrum of cognitive levels and challenges. This diversity ensures that the dataset is robust and suitable for training versatile AI models. Finally, output format and usability were carefully considered by specifying a structured JSON output format with references to the source text. This organization allows for easy integration into downstream AI training and evaluation pipelines, facilitating the efficient use of synthetic data for building and testing AI models in the nuclear domain.

The emphasis on generating a diverse set of questions is essential for developing AI models that are adaptable and capable of handling a wide range of information and queries related to CANDU reactors and the broader nuclear industry. Fundamental recall questions assess basic knowledge and understanding of key concepts, such as "What is the purpose of the calandria in a CANDU reactor?" or "Define the term 'neutron flux' and explain its significance in reactor operation." These questions help reinforce foundational principles and ensure familiarity with essential terminology.

In contrast, technical explanation questions require deeper engagement with the subject matter. The LLM might generate questions like "Describe the process of online refueling in a CANDU reactor, outlining its advantages and disadvantages," or "Explain how the reactor shutdown system (SDS) functions in response to a loss-of-coolant accident." These queries demand a detailed breakdown of operational mechanisms and safety protocols, demonstrating a thorough understanding of CANDU reactor technology.

To assess analytical and problem-solving abilities, the dataset includes multi-step analytical questions that challenge the LLM to evaluate complex scenarios. Examples include "If a control rod fails to insert fully, how would this affect reactor power, and what corrective actions would be necessary?" or "Analyze the impact of a sudden increase in coolant temperature on fuel channel integrity and reactor safety." These questions test the model's ability to reason through intricate operational challenges and provide informed, technical solutions.

By encompassing a wide range of question categories and complexities, the synthetic QnA dataset serves as a rich and diverse training resource for AI models. It enables them to develop a comprehensive understanding of CANDU reactor technology while ensuring they are equipped to address a variety of information needs and challenges in nuclear engineering.

## 5. Evaluation of Synthetic QnA Pairs

This section details the evaluation of the synthetically generated question-answer pairs. We assess the quality of these pairs along several dimensions: semantic diversity, relevance to the source text, and overall question quality. The evaluation leverages embedding analysis and quantitative metrics derived from the generated data.

A few samples of synthetically generated questions and answer pairs are being presented in Table 1 for reference.

**Table 1: QnA Pair Samples**

| Question | Answer |
|---|---|
| Why are Class I power sources essential in a CANDU NPP? | Class I power sources are essential in a CANDU NPP because they provide the necessary DC power to operate critical systems and equipment needed for the safe operation of the nuclear power plant. The loss of Class I power triggers shutdown systems to ensure safety. |
| What voltage levels are used for Class I DC power supply in CANDU plants? | In CANDU plants, the voltage levels used for Class I DC power supply include 48V, 220V/250V, and 400V. |
| Explain the role of Class II power sources in a CANDU NPP. | Class II power sources in a CANDU NPP provide critical AC power derived from Class I DC power sources via inverters. They supply power to systems that can tolerate brief power interruptions and are essential for reactor operation. If Class II power is lost, the reactor will be shut down immediately. |
| What happens in a CANDU NPP if Class II power fails to supply a bus? | If Class II power fails to supply a bus in a CANDU NPP, Class III power sources will be used to support Class II power distribution to ensure continuous power supply and maintain safe operation. |
| What voltage levels are used for Class II AC power supply in CANDU plants? | In CANDU plants, the voltage levels used for Class II AC power supply are 120V and 600V. |

Above QnA Pairs have been synthetically generated from Chapter 11 of the CANDU textbook (Pages 14-15). A critical facet of ensuring the quality of synthetic QnA pairs is human-in-the-loop evaluation, where domain experts meticulously review generated content for accuracy, relevance, and consistency. This approach has long been considered the gold standard for quality assessment in sensitive domains like the nuclear industry. However, despite its effectiveness, human evaluation is highly resource intensive, which can render it impractical for scaling to the vast amounts of data typically. Consequently, while human assessments remain indispensable for verifying the nuanced aspects of synthetic data quality, they must be augmented by automated evaluation techniques. Model-based metrics, semantic similarity analyses, and embedding-based approaches offer complementary, cost-efficient alternatives that can provide continuous, real-time feedback during data generation.

## 5.1 Semantic Diversity

We analyze the semantic diversity of the generated questions by visualizing their embeddings in a lower-dimensional space using t-SNE visualization as shown in Figure 3. In this context, the t-SNE visualization illustrates the semantic distribution of generated and benchmark question embeddings. The orange points represent generated questions, while the blue points represent benchmark questions, providing a comparative reference for evaluating semantic similarity. The benchmark set consists of four MLflow-related queries, which pertain to Machine Learning Operations (MLOps) and have no relation to the nuclear domain. Additionally, a fifth benchmark question on Class IV power in nuclear power plants was included. The visualization highlights that MLflow-related questions appear visually distant from the majority of the generated questions, confirming that they are semantically dissimilar. In contrast, the Class IV power question is positioned closer to the generated question embeddings, indicating its semantic relevance to the nuclear domain. This contrast effectively demonstrates the effectiveness of t-SNE in capturing domain-based differences, providing insights into the diversity and relevance of generated synthetic questions.

## 5.2 Document Relevance

A crucial aspect of question quality is the relevance to the source document chunk. We quantify this relevance using cosine similarity between the question embedding and the corresponding chunk embedding. For each question-chunk pair, we calculate the cosine similarity. The distribution of cosine similarity scores is visualized in a histogram as shown in Figure 4. While most questions exhibit high similarity to their corresponding chunks, we identify questions falling below a threshold of 0.80 for manual inspection. This allows us to pinpoint potential issues where the generated question might not be closely related to the provided text. Manual review of low-similarity questions reveals that, in some instances, the questions are relevant but only pertain to a small portion of the larger text chunk. For instance, a question about circuit breaker components, while relevant to a chunk discussing electrical systems, might have a low similarity score if the chunk primarily focuses on other aspects of the system.

## 5.3 Question Quality Metrics

Beyond relevance, we investigate the overall quality and diversity of the generated questions. We calculate the Shannon entropy of the question set to assess the variety of word usage. Shannon entropy measures the amount of uncertainty or surprise in a message or a set of data. Think of flipping a fair coin. There's a lot of uncertainty about the outcome (heads or tails), so the entropy is high. Now, imagine a coin that always lands on heads. There's no uncertainty at all, so the entropy is zero. In the context of text, entropy measures the diversity of words used. A text with a wide vocabulary and varied sentence structures will have high entropy. A repetitive text with limited vocabulary will have low entropy. The calculated question entropy of 6.63 which was obtained for the generated questions suggests a good level of variation in the questions.

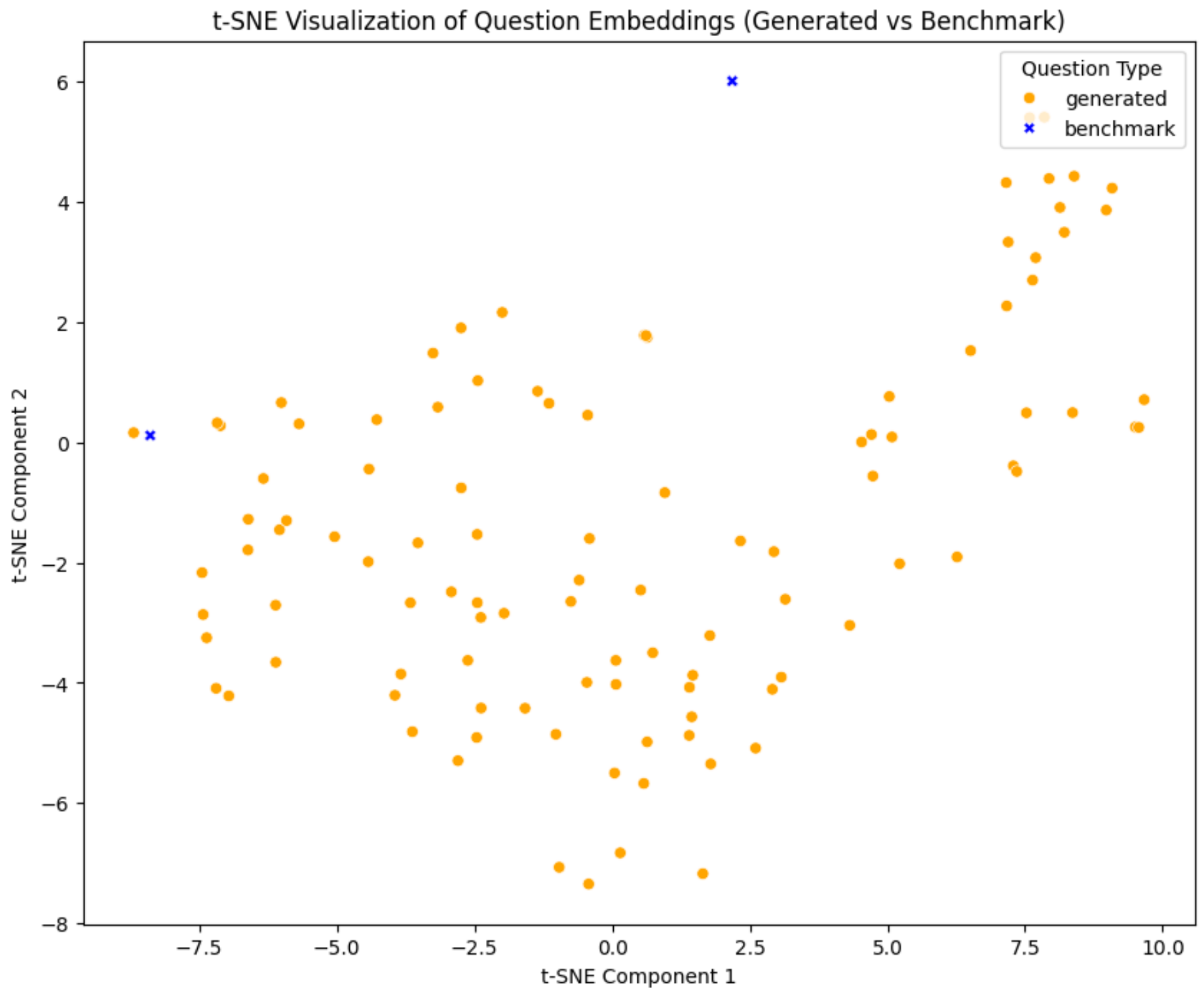

**Figure 3: t-SNE visualization of Questions along with benchmark for comparison**

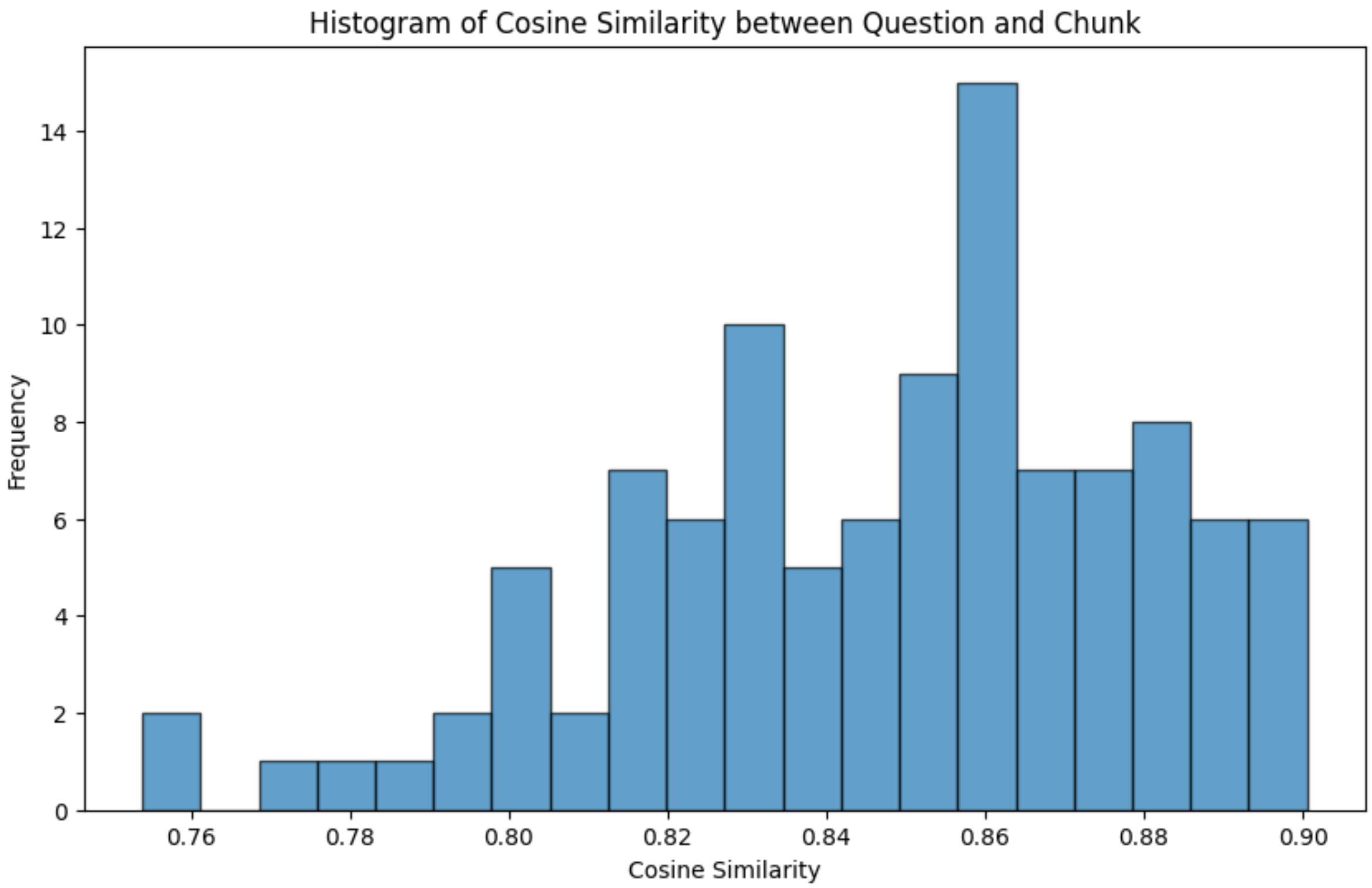

**Figure 4: Histogram for Cosine Similarity between Question and Textbook chunk**

## 6. Future work

Future work should focus on several key areas to enhance scalability, accuracy, and efficiency. Automated synthetic data collection must be prioritized to develop robust pipelines for large-scale gathering and structuring of synthetic nuclear QnA pairs. These pipelines should ensure that the generated data is diverse, domain-specific, and adheres to strict privacy regulations. Additionally, domain-specific fine-tuning and reinforcement learning offer significant potential for improving LLM performance in nuclear applications. Fine-tuning models with synthetic nuclear data, coupled with reinforcement learning techniques, can enhance their ability to handle safety-critical tasks with greater accuracy and reliability.

To maintain high data quality without excessive resource costs, hybrid evaluation strategies should be explored by integrating human-in-the-loop validation with scalable, automated, embedding-based metrics. This approach allows for continuous assessment and refinement of synthetic datasets while minimizing manual workload. Lastly, integration with real-world data presents an opportunity to further improve model proficiency and adaptability. Combining synthetic datasets with limited high-quality real-world nuclear samples can create more robust training frameworks, ensuring that LLMs develop a deeper understanding of nuclear domain knowledge and enhance their applicability to real-world scenarios. By advancing these key areas, future research can significantly improve the reliability and effectiveness of AI applications in the nuclear industry.

## 7. Acknowledgments

This research paper was supported by Ontario Power Generation (OPG) and by The Natural Sciences and Engineering Research Council of Canada (NSERC) and The Canadian Nuclear Safety Commission (CNSC) grant number ALLRP 580442-2022.